\documentclass[runningheads]{llncs}
\usepackage[T1]{fontenc}
\usepackage{subcaption}
\usepackage{caption}
\usepackage{graphicx}
\usepackage{comment} 
\usepackage{hyperref}
\usepackage{todonotes}
\usepackage{amsmath}
\usepackage{amssymb}
\usepackage{mathtools}
\usepackage{booktabs}	
\usepackage{color, colortbl}
\usepackage{bm}
\usepackage{pifont}%
\usepackage{fancyvrb}
\usepackage{backnaur}
\renewcommand{\bnfes}{\epsilon}
\usepackage{multirow}
\usepackage{etoolbox}
\DeclareMathOperator{\edges}{edges}

\usepackage[sort]{cite}
\usepackage{enumitem}
\newlist{questions}{enumerate}{2}
\setlist[questions,1]{label=RQ\arabic*.,ref=RQ\arabic*}
\setlist[questions,2]{label=(\alph*),ref=\thequestionsi(\alph*)}
\makeatletter
\preto{\@verbatim}{\topsep=0pt \partopsep=0pt }
\makeatother

\usepackage{graphicx}
\begin{document}
\title{CompoST: A Benchmark for Analyzing the Ability of LLMs To Compositionally Interpret Questions in a QALD Setting}%
\titlerunning{CompoST: A Benchmark for Compositional Systematicity Testing in QALD}
\author{David Maria Schmidt\inst{1}\orcidID{0000-0001-7728-2884} \and
Raoul Schubert\inst{1}\orcidID{0009-0009-7743-5401} \and
Philipp Cimiano\inst{1}\orcidID{0000-0002-4771-441X}}
\authorrunning{D. M. Schmidt et al.}
\institute{Semantic Computing Group, CITEC, Technical Faculty, Bielefeld University, Bielefeld, Germany\\
\email{\{daschmidt,cimiano\}@techfak.uni-bielefeld.de, raoul.schubert@uni-bielefeld.de}\\}
\maketitle              %
\begin{abstract}
Language interpretation is a compositional process, in which the meaning of more complex linguistic structures is inferred from the meaning of their parts. Large language models  possess remarkable language interpretation capabilities and have been successfully applied to interpret questions by mapping them to SPARQL queries. An open question is how systematic this interpretation process is. Toward this question, in this paper, we propose a benchmark for investigating to what extent the abilities of LLMs to interpret questions are actually compositional. 
For this, we generate three datasets of varying difficulty based on graph patterns in DBpedia, relying on Lemon lexica for verbalization. Our datasets are created in a very controlled fashion in order to test the ability of LLMs to interpret structurally complex questions, given that they have seen the atomic building blocks. This allows us to evaluate to what degree LLMs are able to interpret complex questions for which they ``understand'' the atomic parts. 
We conduct experiments with models of different sizes using both various prompt and few-shot optimization techniques as well as fine-tuning. Our results show that performance in terms of macro $F_1$ degrades from $0.45$ over $0.26$ down to $0.09$ with increasing deviation from the samples optimized on. Even when all necessary information was provided to the model in the input, the $F_1$ scores do not exceed $0.57$ for the dataset of lowest complexity. We thus conclude that LLMs struggle to systematically and compositionally interpret questions and map them into SPARQL queries. 

\end{abstract}
\section{Introduction}
In light of rapidly increasing capabilities of large language models (LLMs), some even proclaiming the ``age of LLMs'' \cite{ageofllms}, the question arises where the limits of current LLMs lie. One central question is whether LLMs are able to truly work and reason in a compositional way. For example, to cite a classic example from Zoltán G. Szabó \cite{comp}, a compositional system understanding both ``brown dog'' and ``black cat'' should understand ``brown cat'' as well. More precisely, Szabó divides compositionality into two sub-properties, namely productivity, describing the ability to produce utterances never encountered before, and systematicity, describing the ability to understand other combinations of known constituents. As figuring out what LLMs have seen during pre-training is notoriously hard if at all possible, in this paper, we focus on a structured investigation of systematicity.

The question of whether LLMs can work in a compositional way has been investigated from different perspectives already, e.g., from a more (complexity-) theoretical point of view \cite{theocomp5,theocomp1,theocomp4,theocomp3,theocomp6}, or with focus on arithmetics \cite{allenaicomp,theocomp2} or logic puzzles \cite{allenaicomp}. Many tasks and works deal with compositionality in one way or another, either directly \cite{allenaicomp,compgap,hupkes_compositionality_2020}, or indirectly by exploring multi-step compositional tasks \cite{zhang_unveiling_2023,nye_show_2021,welleck_naturalprover_2022,saparov_language_2023}. While some of these also investigate question answering \cite{compgap}, the task of mapping natural language questions into SPARQL queries (i.e., Question Answering over Linked Data, QALD) \cite{shekarpour2016question}, to the best of our knowledge, still lacks a structured investigation of the abilities of LLMs to interpret questions in terms of a formal query language in a systematic way. 

As discussed by Dziri et al. \cite{allenaicomp}, many tasks do not require as much compositional behavior as it initially seems, leaving room for overfitting and shortcut learning. This raises the need for specialized datasets, testing the compositional power of LLMs in isolation for different domains. Therefore, in this paper, we present \emph{CompoST}, a \emph{Compositional Systematicity Test} dataset for QALD. In comparison to other datasets, this dataset allows us to test the degree of systematicity of LLM behavior in a controlled setting where irrelevant confounders are removed, so that we can focus on the assessment of the systematicity in isolation from other capabilities. 
In order to create a well-founded dataset, we first operationalize the concept of compositionality w.r.t. \cite{comp} and then generate a corresponding dataset to test the sub-property of systematicity. Finally, we conduct a broad set of zero-shot, few-shot and fine-tuning experiments with this dataset. More precisely, we delve into the following research questions:

\begin{questions}
    \item How can compositionality be operationalized for the question answering over linked data context?\label{rq1}
    \item Do large language models satisfy the defined property of systematicity when using in-context learning?\label{rq2}
    \item Does fine-tuning improve the abilities of large language models to work in a compositional manner?\label{rq3}
\end{questions}

\section{Related Work}

\subsection{Compositional Generalization in NLP}

Compositional generalization in terms of the ability to interpret new combinations of known components has emerged as a critical challenge in evaluating the generalization capacity of neural models. Datasets such as SCAN \cite{lake_generalization_2018}, CFQ \cite{keysers_measuring_2020} and COGS \cite{kim_cogs_2020} have been designed to test this ability under controlled syntactic and semantic conditions: SCAN focuses on mapping simple commands to actions in a synthetic navigation task, CFQ tests generalization over complex SPARQL-like logical forms derived from Freebase and COGS evaluates structural generalization from surface syntax to abstract meaning representations. Unlike QALD tasks, these datasets do not require the generation of executable queries or the linking of expressions to real-world entities. 
In SCAN, Lake and Baroni \cite{lake_generalization_2018} observed that RNN-based models failed to generalize compositionally, succeeding only when test sequences were very similar to those seen during training. Keysers et al. \cite{keysers_measuring_2020} found that Transformer models performed well on CFQ when trained and tested on similar compositions, but accuracy sharply declined when they were tested on structurally novel combinations. Kim and Linzen \cite{kim_cogs_2020} reported that models trained on COGS often failed to generalize across syntactic alternations, such as the dative shift. Furrer et al. \cite{furrer_compositional_2021} showed that even models such as BERT \cite{devlin_bert_2019} underperform on compositional generalization in Freebase-style question answering, and Dessì and Baroni \cite{dessi_cnns_2019} demonstrated similar shortcomings in convolutional architectures on synthetic symbolic tasks. These results suggest that, across architectures, including large-scale Transformer models, there remains a consistent difficulty with systematic generalization to previously unseen combinations of known elements.
Recent work has proposed taxonomies and metrics to distinguish compositional generalization from mere pattern matching \cite{hupkes_compositionality_2020}, which emphasizes the need for benchmarks that isolate compositional competence from surface-level generalization. While this work focuses on a systematicity definition similar to ours, they use an artificial language for a translation task instead of QALD as a task and do not use pre-trained LLMs.

\subsection{Large Language Models and Compositionality}

LLMs have shown impressive zero-shot and few-shot performance across diverse NLP tasks, however their compositional capabilities do still remain contested. Although prompt engineering and in-context learning have extended the apparent flexibility of LLMs \cite{brown_language_2020, wei_chain--thought_2022}, there are still limitations in their ability to generalize from known linguistic or logical primitives to new combinations \cite{herzig_unlocking_2021, piantadosi_meaning_2022}. For instance, Petty et al. \cite{petty_impact_2024} have found that increasing transformer depth yields diminishing returns for compositional generalization tasks. In a similar vein, Yang et al. \cite{yang_exploring_2024} have observed that training LLMs on higher-order compositional instructions improves performance on simpler tasks, but not vice versa. Furthermore, Ismayilzada et al. \cite{ismayilzada_evaluating_2025} have reported that instruction-fine-tuned multilingual models struggle when it comes to morphological compositional generalization in agglutinative languages such as Turkish and Finnish, particularly when applied to novel word roots. These findings suggest a discrepancy between LLMs' surface-level fluency and their deeper semantic systematicity. 

Dziri et al. \cite{allenaicomp} investigate the limits of Transformer-based architectures on compositional reasoning tasks, e.g., multiplication and logic puzzles. They demonstrate that even advanced LLMs exhibit systematic failures when faced with reasoning chains that are beyond the complexity seen during training. Although Dziri et al. also investigate the compositional abilities of LLMs, they focus on different tasks, use a set of older and less diverse models as well as a more implicit compositionality definition. Press et al. \cite{compgap} propose an empirical framework for quantifying what they call the ``compositionality gap'' in language models. By introducing diagnostic datasets and evaluation metrics, they showcase how current models often rely on statistical heuristics rather than exhibiting true compositional understanding. They also test question answering as a task, although not QALD. Moreover, the problems tested in \cite{compgap} are limited in size and complexity by mostly focusing on $2$-hop question answering, together with older and less diverse tested LLMs. In contrast, our approach generalizes from that $2$-hop setting to a general sub-graph-based notion for determining the parts of a question that allows to easily scale the complexity of the benchmarks. 

\subsection{Compositionality in QALD}

QALD represents a natural way of testing for compositional generalization, as building SPARQL queries involves, among other challenges, logical composition, nesting relations, and linking entities. Previous approaches in semantic parsing have addressed these challenges by explicitly modeling the structure of SPARQL queries derived from natural language \cite{yih_semantic_2015, ghidini_lc-quad_2019, qald10, qald9plus, qald9}. However, these systems typically require domain-specific grammars or supervision, which greatly limits their scalability.
Recent investigations have focused on the capacity of LLMs to answer knowledge base questions without explicit parsing. In contrast, Schmidt et al. \cite{neodudes,neodudesPoster} aim to combine the strengths of both LLMs as well as symbolic approaches in a compositional QALD pipeline. Their system works by utilizing lexical entries and explicit meaning representations together with neural parts for, e.g., dependency parsing or selecting the most promising SPARQL query. 

While LLMs perform reasonably on atomic questions (i.e., involving a single RDF triple), they do still struggle with multi-hop or nested constructions \cite{he_mintqa_2025, chakraborty_multi-hop_2024, panda_holmes_2024}, which are prevalent in QALD. Recent studies suggest that these failures may stem from LLMs' limited ability to systematically map natural language inputs to the compositional structures that are required for accurate knowledge base reasoning. For instance, Chen et al. \cite{theocomp1} demonstrate that integrating explicit knowledge graphs enhances LLM performance on multi-hop question answering tasks, at the same time indicating that LLMs alone struggle with such compositional reasoning. In a similar vein, Zhao et al. \cite{zhao_exploring_2024} find that LLMs possess the necessary knowledge components but fail to combine them effectively in novel contexts, pointing out a deficiency in systematic compositionality.
As such, this paper positions itself at the intersection of these concerns: operationalizing compositionality through limited linguistic units, DBpedia-based property verbalizations, and testing whether LLMs exhibit systematic generalization w.r.t. QALD. By grounding the experimental design in a formal notion of compositionality and providing the CompoST dataset, our aim is to offer a controlled, yet scalable, approach to evaluating this dimension of model behavior.

\section{Methods}

\subsection{Operationalization of Compositionality}\label{sec:operationalization}

For our definition of compositionality in a QALD context, we rely on work by Zoltán G. Szabó~\cite{comp}, where compositionality is described as consisting of two sub-properties, namely productivity and systematicity. We slightly generalize the definitions and arguments made in that work to also include generative LLMs. These definitions are also similar to the systematicity and productivity tests described in \cite{hupkes_compositionality_2020}. The definitions used in this work are, therefore, as follows:

\begin{definition}[Productivity \cite{comp}]\label{def:prod}
    The property of productivity is satisfied iff an agent can understand\footnote{As the original definition refers to ``understand'' in a human context, we use the following working definition of understanding for generative LLMs in this paper: An LLM understands a sentence w.r.t. some knowledge graph iff it can generate a correct SPARQL query for that respective sentence and knowledge graph.} complex expressions it never encountered before.
\end{definition}

\begin{definition}[Systematicity \cite{comp}]\label{def:sys}
    The property of systematicity is satisfied iff an agent that understands a number of complex expressions $e_1, \dots, e_n$ also understands all other complex expressions that can be built up from the constituents of $e_1, \dots, e_n$ using syntactic rules employed in building up their structures.
\end{definition}

As we want to operationalize compositionality and develop a benchmark dataset for LLMs, we need to think of ways to test these definitions. For Def. \ref{def:prod}, an obvious challenge in applying it to LLMs is that i) for many models, the datasets used for pre-training are not publicly available, and ii) even if they are, e.g., for models like OLMo \cite{olmo}, they are huge and it is hardly verifiable whether an expression is or is not part of the training data beyond exact string matches. We therefore focus on the systematicity property as stated in Def. \ref{def:sys} only.

In order to operationalize Def. \ref{def:sys}, there are multiple ways in which the term ``constituents'' can be interpreted in the QALD context. On the one hand, this could be interpreted as constituents w.r.t. natural language, i.e., words or sentence structures, and, on the other hand, w.r.t. SPARQL, syntax elements like triple patterns. In this paper, we focus on investigating the behavior of LLMs w.r.t. SPARQL constituents and verbalize the corresponding SPARQL queries with only slight linguistic variations. In this way, we want to test the basic abilities of LLMs to work in a compositional, i.e., systematic, manner, without having to generalize over many  potential verbalizations. 

\subsection{CompoST Dataset Creation}

For creating a dataset to test the systematicity property, we thus need pairs of questions and SPARQL queries (named ``benchmark item'' or just ``item'' in the following) such that we know that the sufficiency criterion is fulfilled, i.e.,  we know whether or not a set of benchmark items contains all necessary constituents to build another specific item. This requirement also prevents us from using existing QALD datasets. QALD benchmarks typically consist of a large number of mostly different questions, but lack two aspects that are necessary to test compositional systematicity. More precisely, they a) usually do not have large numbers of different combinations of the same ``basic building blocks'' and b) do not have known relations between two benchmark items that would allow us to determine whether the sufficiency criterion is satisfied or not. Thus, we create a new dataset that meets these requirements. For our dataset, this is achieved by first generating graph patterns of a certain size and structure, then fetching instances of that pattern from a DBpedia snapshot from December 2022, and finally verbalizing the full pattern as well as all connected sub-patterns in the same structured fashion. For each resulting set of benchmark items, we know whether the sufficiency criterion is satisfied if the graph patterns connected to a set of items are an (exact) edge set cover of the considered other item's graph pattern.
More precisely, we use ``pitchfork-like''/star patterns of variable depth and breadth. An instance for depth $3$ and breadth $3$ is given in Fig. \ref{fig:3x3inst}. These pattern instances as well as all connected sub-graphs of those patterns form the basis of our benchmark dataset. 
\begin{figure}[t]
    \centering
    \resizebox{0.65\width}{!}{%
    \begin{tikzpicture}[node distance=3.5cm, align=center]%
        \node (res) {dbr:Angela\_Merkel};        
        \node (o12) [above of=res] {dbr:Mali\_War};
        \node (o11) [right of=o12] {dbr:Operation\_Freedom's\_\\Sentinel};
        \node (o13) [left of=o12] {dbr:Operation\_Enduring\_\\Freedom\_–\_Horn\_of\_Africa};
        \draw[->] (o11) -- node[left,above, sloped]{dbo:commander} (res);
        \draw[->] (o12) -- node[left,above, sloped]{dbo:commander} (res);
        \draw[->] (o13) -- node[left,above, sloped]{dbo:commander} (res);
        \node (o22) [above of=o12] {dbr:Ba\_Ag\_Moussa};
        \node (o21) [right of=o22] {dbr:United\_States\_\\Army\_Central};
        \node (o23) [left of=o22] {dbr:United\_States\_Navy};
        \draw[->] (o21) -- node[left,above, sloped]{dbo:battle} (o11);
        \draw[->] (o22) -- node[left,above, sloped]{dbo:battle} (o12);
        \draw[->] (o23) -- node[left,above, sloped]{dbo:battle} (o13);
        \node (o32) [above of=o22] {dbr:Mali\_War};
        \node (o31) [right of=o32] {dbr:Battle\_of\_Fort\_Driant};
        \node (o33) [left of=o32] {dbr:Thomas\_C.\_Butler};
        \draw[->] (o31) -- node[left,above, sloped]{dbo:commander} (o21);
        \draw[->] (o32) -- node[left,above, sloped]{dbo:commander} (o22);
        \draw[->] (o33) -- node[left,above, sloped]{dbo:almaMater} (o23);
    \end{tikzpicture}}
    \caption{RDF graph of SPARQL graph pattern instance with depth of $3$ and breadth of $3$ with concrete entities and properties from DBpedia.}%
    \label{fig:3x3inst}
\end{figure}
However, the pattern instance of Fig. \ref{fig:3x3inst} still needs to be translated into SPARQL queries and corresponding verbalizations. For this, we rely on $1070$ hand-crafted Lemon lexical entries \cite{lemon} for $147$ DBpedia properties, roughly corresponding to the properties covered in QALD-9 \cite{qald9}. Of these entries, $245$ are \texttt{NounPPFrame} entries for verbalizing DBpedia properties and $825$ are about verbalizing different \texttt{rdf:type}, \texttt{dbo:country}, \texttt{dbo:nationality} and \texttt{dbo:industry} patterns, e.g., ``\texttt{?v dbo:nationality dbr:Germany}'' as ``is German''. For this, we reused and extended the lexicon of Schmidt et al. \cite{neodudes,neodudesPoster} as well as parts of the code for parsing and using the lexical entries.

\begin{figure}[t]
    \scriptsize
    \begin{enumerate}
        \item Question: Who is the spouse of Michelle Obama and parent of Malia Obama?
    
        SPARQL: \texttt{dbr:Michelle\_Obama dbo:spouse ?result. dbr:Malia\_Obama dbo:parent ?result.}
        \item Question: Who is the spouse of the child of Marian Shields Robinson?

        SPARQL: \texttt{dbr:Marian\_Shields\_Robinson dbo:child ?v1. ?v1 dbo:spouse ?result.}
    \end{enumerate}
    Question: Who is the spouse of the child of Marian Shields Robinson and parent of Malia Obama?

    SPARQL: \texttt{dbr:Marian\_Shields\_Robinson dbo:child ?v1.} \texttt{?v1 dbo:spouse ?result.} \texttt{dbr:Malia\_Obama dbo:parent ?result.}
    
    \caption{Example question-query-pairs. $1$ and $2$ together contain all relevant information to compose the bottom query. Only triple patterns of queries shown.}
    \label{fig:exampleset}
\end{figure}

A Lemon lexical entry consists of a canonical form of the respective word or phrase, the corresponding preposition (if applicable) as well as the associated DBpedia property, among other things. Based on this information, we can verbalize a given ``pitchfork-like'' graph pattern such that we ask for \texttt{?result}. Examples are given in Fig. \ref{fig:exampleset}, illustrating how a set of items can contain all necessary information to compose another item. By using the information from $1$ and $2$, a compositional system should be able to generate the bottom query. %

As the structure of these questions and queries is quite simple and monotonous, we additionally increased the diversity of our dataset in two ways: i) inverting the direction of some triples in the pattern and ii) adding verbalizations of \texttt{rdf:type} or, e.g., \texttt{dbo:nationality} values to the query. For further details about the dataset generation process, we published the corresponding generation code in our software artifact.
All in all, the generated questions can be described by the grammar in Extended Backus-Naur Form (EBNF, \cite{bnf}) presented in Fig. \ref{fig:grammar}. %
\begin{figure}[t]
    \centering
    \scriptsize
    \begin{bnf*}
      \bnfprod{S}{\bnfts{Give me} \bnfsp \bnfpn{DP} \bnfsp (\bnfpn{CP} \bnfor \bnfes) \bnfsp \bnfpn{PP} \bnfsp \bnfts{.}}\\
      \bnfmore{\bnfor \bnfts{Who is} \bnfsp \bnfpn{DP} \bnfsp (\bnfpn{CP} \bnfor \bnfes) \bnfsp \bnfpn{PP} \bnfsp \bnfts{?}}\\
      \bnfprod{VP}{ \bnfts{is} \bnfsp \bnfpn{DP}}\\
      \bnfmore{\bnfor \bnfts{works} \bnfsp \bnfpn{PP}}\\
      \bnfmore{\bnfor \bnfpn{VP} \bnfsp \bnfpn{CP}}\\
      \bnfmore{ \bnfor (\bnfpn{VP} \bnfor \bnfpn{DP}) \bnfsp \bnfts{and} \bnfsp (\bnfpn{VP} \bnfor \bnfpn{DP})}\\
      \bnfprod{DP}{\bnfpn{Det} \bnfsp \bnfpn{N} \bnfor \bnfpn{Det} \bnfsp \bnfpn{N} \bnfsp \bnfts{industry}}\\
      \bnfprod{Det}{\bnfts{the} \bnfor \bnfts{a} \bnfor \bnfts{an} \bnfor \bnfes}\\
      \bnfprod{CP}{\bnfts{, that} \bnfsp \bnfpn{VP} \bnfsp \bnfts{,}}\\
      \bnfprod{PP}{\bnfpn{P} \bnfsp \bnfpn{DP} \bnfor \bnfpn{PP} \bnfsp \bnfpn{CP} \bnfor \bnfpn{PP} \bnfsp  \bnfpn{PP} \bnfor \bnfpn{PP} \bnfsp \bnfts{and} \bnfsp \bnfpn{PP}}\\
      \bnfprod{N}{\bnftd{canonical form of lexical entry}}\\
      \bnfmore{\bnfor \bnftd{label of corresponding DBpedia entity or literal value}}\\  
      \bnfprod{P}{\bnftd{preposition of NounPPFrame lexical entry} \bnfor \bnfts{in}}  
    \end{bnf*}%
    \caption{EBNF \cite{bnf} of the sentences generated for the CompoST dataset.}%
    \label{fig:grammar}%
\end{figure}%
However, that grammar generates a super-set of the sentences that can actually be generated by our approach. Therefore, some additional constraints enforced during the dataset generation are omitted. For example, the choice of ``works in ...'', ``is a'', ``is an'' or just ``is'' depends on the respective canonical form, the corresponding property, as well as the specific lexical entry type. %

For the final dataset, patterns with the following depth and breadth combinations of the basic pattern were used: breadth $2$ \& depth $2$, breadth $2$ \& depth $3$, breadth $3$ \& depth $2$, and breadth $3$ \& depth $3$. For each, we searched for matching pattern instances in DBpedia and, based on the instances found, generated a verbalization of those base pattern instances as well as of all connected sub-patterns of the respective instances. In cases of multiple available lexical entries for a property, we chose the entries consistently between the base pattern and its sub-patterns by seeding the random choices accordingly. Similarly, we fetched all \texttt{rdf:type}, \texttt{dbo:country}, \texttt{dbo:nationality} and \texttt{dbo:industry} objects for all entities of the instance and selected a random sample of random size 
of the available objects for which a corresponding lexical entry was available. This sample was then fixed for that pattern instance and verbalized as a ``that'' clause for all (sub-)patterns that include the respective node at an appropriate position. All those choices are necessary in order to ensure that, as Def. \ref{def:sys} states, all necessary constituents are present in the sub-patterns such that one should be able to compose those parts to verbalizations of larger (sub-)patterns. %

The generated instances, their verbalizations and the corresponding SPARQL queries were then split equally among three datasets of varying difficulty, with every dataset having in total $75$ base pattern instances as well as the corresponding sub-graphs. In total, this makes $2803$ question-query-pairs per dataset. The splitting was done based on the portion of the number of pattern edges (thus, ``that'' clause edges were excluded from that calculation) up to which sub-patterns were added to the training split. More precisely, they were split into \emph{easy} (sub-patterns with up to $75\%$ of the total base pattern edges were included in the train split, e.g., for breadth $3$ \& depth $3$ with a total of $9$ edges, patterns with up to $6$ edges were included in train), \emph{medium} ($50\%$) and \emph{hard} ($25\%$, with an absolute minimum of two edges to cover all phenomena, such as chaining triples or branching with ``and''). For the validation and test splits, we chose the lower and upper half of the remaining data sorted by the number of base pattern edges, respectively.

Additionally, based on these datasets, we created additional variants where an edge set cover of the target pattern, i.e., all information necessary to compose the correct answer, is given in the input. In the following, these variations will be called \emph{self-contained tasks}, as no knowledge beyond the input is necessary to solve it. Analogously, the original tasks are called \emph{classic} tasks. The self-contained tasks were created by taking the validation items for the new train and the test items for the new test datasets together with an edge set cover of the target pattern taken from the train dataset, comprising up to $10$ examples\footnote{In some cases, there were less than $10$ smaller samples related to the target pattern and thus given in the input. For further details, please refer to our software artifact.}. 

With these datasets, which together comprise CompoST, we investigate the ability of models to generalize beyond what they have seen either during prompt optimization, in the given examples or during fine-tuning. A system which is able to truly generalize in a compositional way would be expected to show almost no performance degradation, as all necessary information is already in sub-patterns with as few as one triple, or two triples to cover combination phenomena.

\section{Experiments}\label{sec:experiments}
In the following, we describe the experiments conducted with CompoST.

\paragraph{Zero-Shot Prompting:}
 As a first baseline, we consider zero-shot prompting. In order to run the experiments in a structured manner together with, e.g., prompt optimization techniques, we use the DSPy framework \cite{dspy1,dspy2} for our in-context learning experiments. We evaluate a broad set of models, namely Llama 3.3 ($70$B) \cite{llama}, Phi-4 ($14$B) \cite{phi4}, Qwen2.5-Coder ($7$B) \cite{qwen1,qwen2}, OLMo 2 ($7$B) \cite{olmo} and GPT-4o-mini \cite{gpt4}. As optimization techniques, we used none, COPRO, MIPROv2 (light, medium and heavy) together with plain as well as chain of thought prompting.
In each experiment, the model is given a question from the dataset and expected to generate a corresponding SPARQL query for DBpedia. Finally, the results of both queries are compared and the $F_1$ scores are calculated. The optimization strategies are given the training split and can optimize the prompt based on an evaluation method that calculates the $F_1$ score for a predicted query compared to the gold standard.

\paragraph{Few-Shot Prompting:}

As generating SPARQL queries for natural language questions is a hard task, one might argue that, besides prompt optimization, examples are necessary for the model to properly understand the task. Therefore, we use MIPROv2 (light, medium and heavy), LabeledFewShot, BootstrapFewShot and BootstrapFewShotWithRandomSearch choosing the shots given to the model, as well as (in case of MIPROv2) optimizing the prompt at the same time, again paired with plain or chain of thought prompting. 
In these experiments, the models are additionally given a number of examples, i.e., ``shots'', chosen by the respective optimization strategies. For choosing those shots, the strategies are also given the SPARQL queries for the train split and have the same information available as the zero-shot experiments.
For the self-contained tasks, we evaluated MIPROv2 (light, medium and heavy) with and without chain of thought prompting, as the given shots are a fixed part of the input already.

\paragraph{Fine-Tuning:}

For particularly hard tasks, even with the most recent models, it might be necessary to apply fine-tuning when in-context learning fails. Therefore, we also conducted experiments with fine-tuned models. In particular, we fine-tuned GPT-4o-mini, OLMo 2 and Qwen2.5-Coder. Although hardware limitations were no issue for GPT-4o-mini fine-tuning because of the OpenAI API usage, we chose the two $7$B parameter models OLMo 2 and Qwen2.5-Coder over Llama 3.3 and Phi-4 as this was the largest model size we could fine-tune in reasonable time with our available hardware. %
For the classic tasks, these models were fine-tuned on a zero-shot setting, i.e., given a basic task description \footnote{\scriptsize \texttt{Given a natural language question about a specific entity, generate a SPARQL query that retrieves relevant information from DBpedia.}} together with the input question and fine-tuned to output just the SPARQL query. For self-contained tasks, we fine-tuned the best-performing model of the classic tasks, namely GPT-4o-mini. In this setting, the model was additionally given the relevant edge set cover of example items. Thus, the model is fine-tuned to compose the answer based on question-query-pairs that are given to the model.

\subsection{Experimental Settings}\label{sec:expsettings}

For all in-context learning experiments, we used the OpenAI API for GPT-4o-mini, evaluating the \texttt{gpt-\allowbreak 4o-\allowbreak mini-\allowbreak 2024-07-18} version, and Ollama\footnote{\scriptsize \url{https://ollama.com/}} to serve all other models. More precisely, we used the default Ollama models with \texttt{Q4\_K\_M} quantization, i.e., Llama 3.3 with $70$B parameters, Phi-4 with $14$B parameters, and Qwen2.5-Coder with $7$B parameters. Additionally, we used the \texttt{7b-\allowbreak 1124-\allowbreak instruct-\allowbreak q4\_K\_M} version of OLMo 2. The models used for (classic) fine-tuning were GPT-4o-mini 
as well as Qwen2.5-Coder (\texttt{Qwen/\allowbreak Qwen2.5-\allowbreak Coder-\allowbreak 7B-\allowbreak In\allowbreak struct}) and OLMo 2 (\texttt{allenai/\allowbreak OLMo-2-\allowbreak 1124-\allowbreak 7B-\allowbreak Instruct}). For fine-tuning GPT-4o-mini, we used the OpenAI API, leaving all parameters to ``auto''. For the other models, we performed a hyperparameter search 
using Optuna \cite{optuna} and fine-tuned the models using PyTorch Lightning \cite{lightning} with DeepSpeed \cite{deepspeed}. For our experiments, we had up to $10$ Nvidia A40 GPUs available, utilizing nodes with up to four A40 GPUs paired with $120$ CPU cores and $443$GB of RAM. With quantization and DeepSpeed \cite{deepspeed}, we were able to fine-tune the given $7$B models on a single node. The prompting experiments not using the OpenAI API were conducted with Ollama instances running on a single A40 GPU each, grouping experiments by models accordingly to avoid reloading.

To measure the ``raw'' performance, we get the results for both the gold standard, as well as the generated SPARQL query of a DBpedia snapshot from December 2022\footnote{\scriptsize \url{https://databus.dbpedia.org/dbpedia/collections/dbpedia-snapshot-2022-12}}. We then compare the two result sets and calculate true positives, false positives, etc. in the standard way. Based on the confusion matrix, we can then calculate $F_1$ scores at a macro level (calculating $F_1$ score per question and using the mean of the respective scores as the final score). The $F_1$ scores for single questions are available to the DSPy \cite{dspy1,dspy2} optimizers.

Additionally, we calculate a second set of scores focusing more on the compositionality aspect, evaluating whether or not the model behaves in a compositional way. For this, we define the confusion matrix values w.r.t. expectations of compositional behavior. More precisely, this means that true positives (TP) are those correctly-answered benchmark items for which all sub-problems, i.e., all sub-patterns of the respective target graph pattern, are correctly answered as well. False positives (FP) are all other correctly-answered tasks, i.e., where one or more sub-problems are wrong. Finally, false negatives (FN) are those incorrect answers for which an edge set cover of the target graph is correctly answered, but the actual question is still answered incorrectly. %
\begin{align*}%
    &\mathcal{I} = \mathcal{C} \cup \mathcal{W} %
    &\mathit{TP} = \{i \in \mathcal{C} \mid \forall i' \in \mathcal{I}\colon i' \subseteq i \Rightarrow i' \in \mathcal{C}\ldotp\}
    &&\mathit{FP} = \mathcal{C} \setminus \mathit{TP}
\end{align*}\vspace{-6mm}
\[\mathit{FN} = \{i \in \mathcal{W} \mid \exists\mkern2mu\alpha \subseteq \mathcal{C}\colon \bigcup\limits_{i' \in \alpha} \edges(i') = \edges(i)~\land~\forall i' \in \alpha\colon i' \subset i \ldotp\ldotp\}\]%
with $\mathcal{I}$ being the set of all benchmark items (represented by their graph pattern instance), $\mathcal{C}$ the set of correctly-answered items and $\mathcal{W}$ the set of items with a wrong answer. Furthermore, $\edges(i \in \mathcal{I})$ returns the set of edges of the graph pattern instance of a benchmark item. Based on this, the compositionality-adjusted $F_1$ score is then calculated per question and aggregated as a macro $F_1$ score for the resulting compositionality $F_1$ score displayed in Table \ref{tab:results}.

\section{Results}
As we conducted over $400$ experiments in total, this section will only discuss the results of the best-performing models (w.r.t. overall macro $F_1$ score) and optimization strategies per category. %
\begin{table}[t]%
\caption{Performance of the best-performing model and optimization strategy for each category. * = self-contained tasks, BFRS = BootstrapFewShotWithRandomSearch, CoT = Chain of Thought, MIPRO: H = Heavy, M = Medium}\label{tab:results}
\centering
\begin{tabular}{r|c|c|ccc|ccc}
   \toprule
         Experiment &  \multicolumn{2}{c|}{Best Configuration} & \multicolumn{3}{c|}{Macro $F_1$} & \multicolumn{3}{c}{Compositionality $F_1$}  \\
                    &    Model    &        Optimization        & Train  & Validation &    Test    & Train  & Validation &         Test          \\ \midrule
                                                         \multicolumn{9}{c}{Easy (75\%)}                                                        \\ \midrule
          zero-shot &  Llama 3.3  & $\mathit{MIPRO}_{H + CoT}$ & $0.17$ &   $0.04$   &   $0.01$   & $0.10$ &   $0.00$   &        $0.00$         \\
           few-shot &  Llama 3.3  & $\mathit{MIPRO}_{M + CoT}$ & $0.23$ &   $0.12$   &   $0.09$   & $0.15$ &   $0.00$   &        $0.00$         \\
         fine-tuned & GPT-4o-mini &        fine-tuning         & $0.92$ &   $0.58$   &   $0.45$   & $0.79$ &   $0.49$   &        $0.31$         \\
          few-shot* &  Llama 3.3  &    $\mathit{MIPRO}_{M}$    & $0.56$ &            &   $0.57$   & $0.72$ &            &        $0.52$         \\ 
        fine-tuned* & GPT-4o-mini &        fine-tuning         & $0.72$ &            &   $0.48$   & $0.84$ &            &        $0.53$         \\ \midrule
                                                        \multicolumn{9}{c}{Medium (50\%)}                                                       \\ \midrule
          zero-shot &  Llama 3.3  &    $\mathit{MIPRO}_{H}$    & $0.19$ &   $0.03$   &   $0.01$   & $0.14$ &   $0.00$   &        $0.16$         \\
           few-shot &  Llama 3.3  &   $\mathit{BFRS}_{CoT}$    & $0.29$ &   $0.15$   &   $0.09$   & $0.18$ &   $0.02$   &        $0.03$         \\
         fine-tuned & GPT-4o-mini &        fine-tuning         & $0.94$ &   $0.57$   &   $0.26$   & $0.78$ &   $0.60$   &        $0.22$         \\
          few-shot* &  Llama 3.3  &    $\mathit{MIPRO}_{H}$    & $0.68$ &            &   $0.48$   & $0.76$ &            &        $0.33$         \\ 
        fine-tuned* & GPT-4o-mini &        fine-tuning         & $0.92$ &            &   $0.55$   & $0.95$ &            &        $0.57$         \\ \midrule
                                                         \multicolumn{9}{c}{Hard (25\%)}                                                        \\ \midrule
          zero-shot &  Llama 3.3  & $\mathit{MIPRO}_{H + CoT}$ & $0.25$ &   $0.07$   &   $0.04$   & $0.08$ &   $0.18$   &        $0.06$         \\
           few-shot & GPT-4o-mini & $\mathit{MIPRO}_{M + CoT}$ & $0.36$ &   $0.13$   &   $0.08$   & $0.11$ &   $0.09$   &        $0.07$         \\
         fine-tuned & GPT-4o-mini &        fine-tuning         & $0.95$ &   $0.34$   &   $0.09$   & $0.61$ &   $0.45$   &        $0.10$         \\
          few-shot* &  Llama 3.3  &    $\mathit{MIPRO}_{M}$    & $0.89$ &            &   $0.51$   & $0.92$ &            &        $0.43$         \\ 
        fine-tuned* & GPT-4o-mini &        fine-tuning         & $0.99$ &            &   $0.30$   & $0.99$ &            &        $0.41$         
\end{tabular}
\end{table}
As shown in Table \ref{tab:results}, the best-performing models (in terms of macro $F_1$ score) are either Llama 3.3 or GPT-4o-mini across all categories and datasets. Additionally, for in-context learning, MIPROv2 prompt optimization works best in most cases, with \emph{medium} few-shot being the only exception, achieving the best results for BootstrapFewShotWithRandomSearch. However, fine-tuning outperforms in-context learning in all cases in terms of macro $F_1$ scores for the classic tasks. For the self-contained task variants, it is slightly different. Here, few-shot experiments outperform the fine-tuned approaches for the \emph{easy} ($0.57$ vs. $0.48$) and \emph{hard} ($0.51$ vs. $0.30$) datasets. Considering the different datasets with different portions of the total edge count being present in training data, it can generally be observed that the models perform worse the more the tasks differ from their training samples. This effect is especially strong for the fine-tuned models, with, e.g., fine-tuned test macro $F_1$ scores degrading from $0.45$ for \emph{easy} over $0.29$ for \emph{medium} to $0.09$ for \emph{hard}. Notably, in all datasets, all relevant information is part of the training data. Thus, only the number and complexity of components that need to be composed differs.

For the compositionality $F_1$ score, all classic configurations show poor performance. No approach achieves a higher test score than $0.31$, which even degrades to $0.22$ for \emph{medium} and $0.10$ for \emph{hard}. But also for self-contained configurations, where the models are optimized for composing the pieces already present in the input, the compositionality $F_1$ score does not exceed $0.57$ for the test split, with just $0.43$ being the best result for the hard dataset. Comparing in-context learning and fine-tuning, no clear trend is observable. For the \emph{easy} dataset, the fine-tuned approach's compositionality $F_1$ score is slightly higher than for the best few-shot approach ($0.53$ vs. $0.52$), although the macro $F_1$ score is lower ($0.48$ vs. $0.57$). In contrast, the opposite is true for the \emph{medium} and \emph{hard} datasets.%

\begin{figure}[t]
    \centering
    \begin{subfigure}{0.49\textwidth}
    \centering
    \resizebox{0.9\textwidth}{!}{\includegraphics{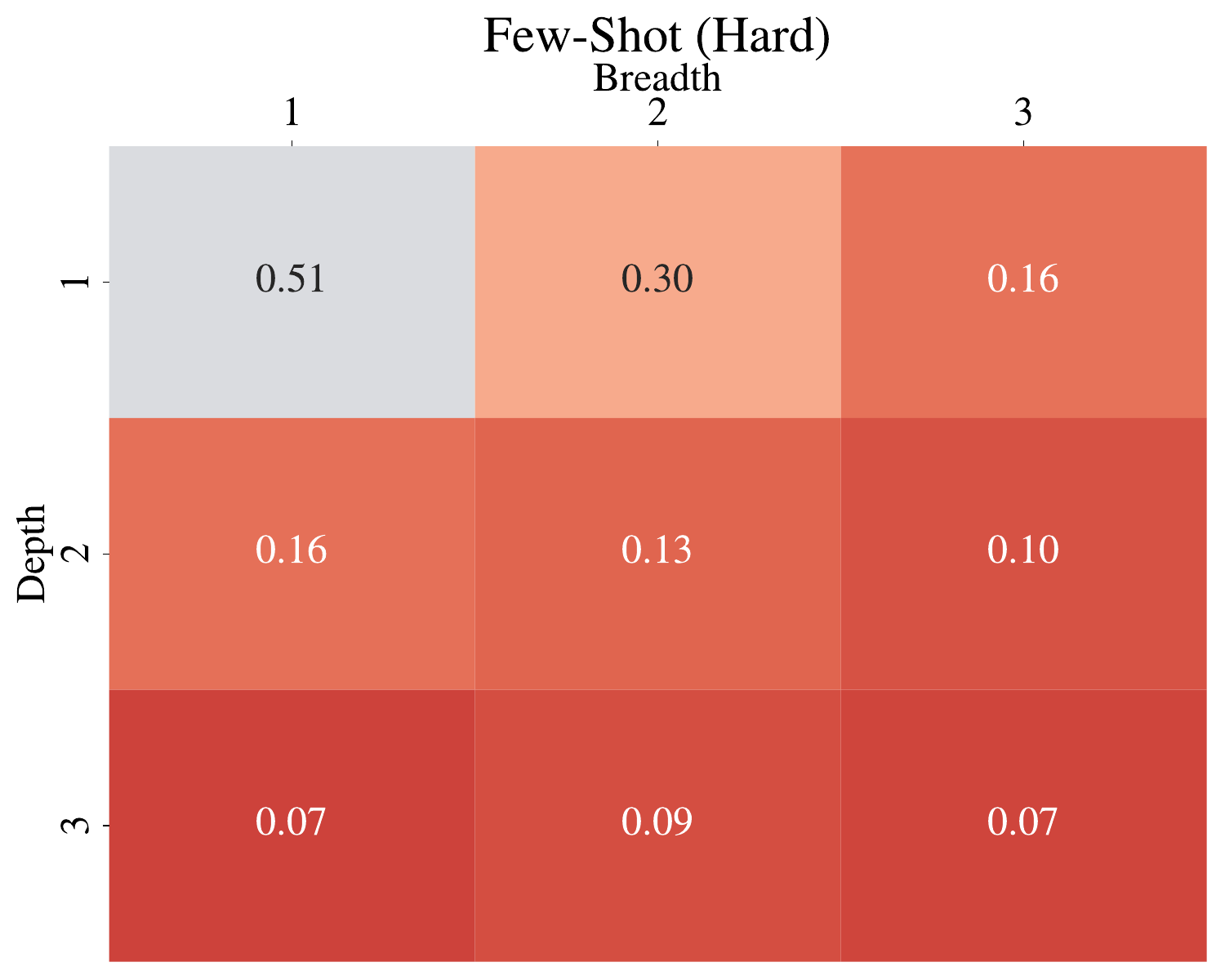}}
    \caption{Best Few-Shot Approach}
    \label{fig:heatmap-best-max-few}
    \end{subfigure}
    \begin{subfigure}{0.49\textwidth}
    \centering
    \resizebox{0.9\textwidth}{!}{\includegraphics{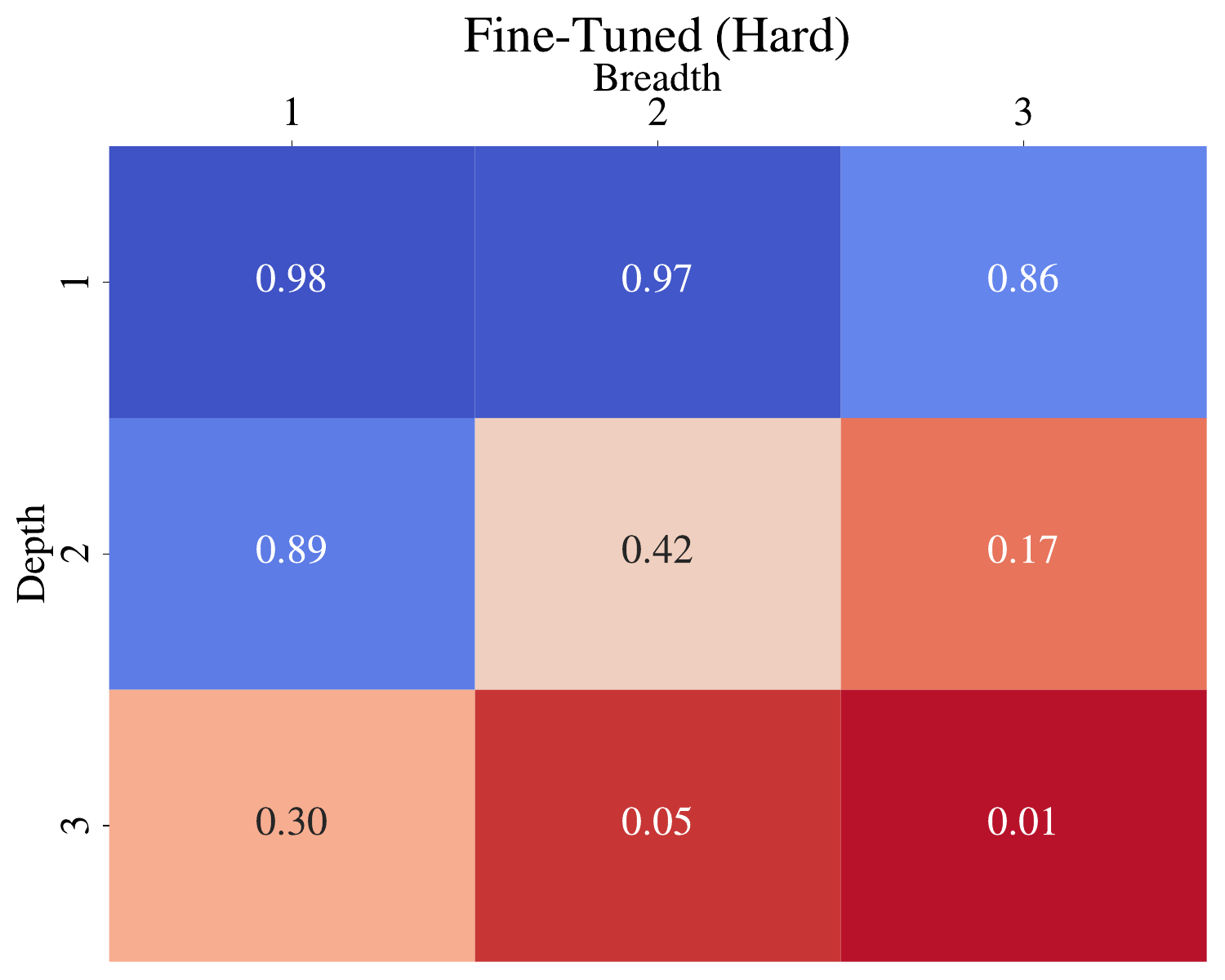}}
    \caption{Best Fine-Tuned Approach}
    \label{fig:heatmap-best-max-ft}
    \end{subfigure}
    \caption{Macro $F_1$ scores of best approaches of the respective category for the \emph{hard} dataset, grouped by graph pattern depth and breadth. Training data comprised samples up to two edges, i.e., up to breadth $2$, depth $1$ and depth $2$, breadth $1$.}
    \label{fig:heatmap-best-max}
\end{figure}

The performance of the models depending on the depth and breadth of the given question and respective SPARQL query is further examined in Fig. \ref{fig:heatmap-best-max} by the example of the hard dataset. As we observed in Fig. \ref{fig:heatmap-best-max-few}, the single-triple queries work comparably well, whereas increasing depth and breadth substantially hurts performance. The fine-tuned approach, see Fig. \ref{fig:heatmap-best-max-ft}, works almost perfectly for the training data and successfully generalizes to breadth $3$, however, increasing both depth and breadth degrades performance down to $0.01$ for depth and breadth $3$.

\begin{figure}[t]
    \centering
    \resizebox{0.99\textwidth}{!}{\includegraphics{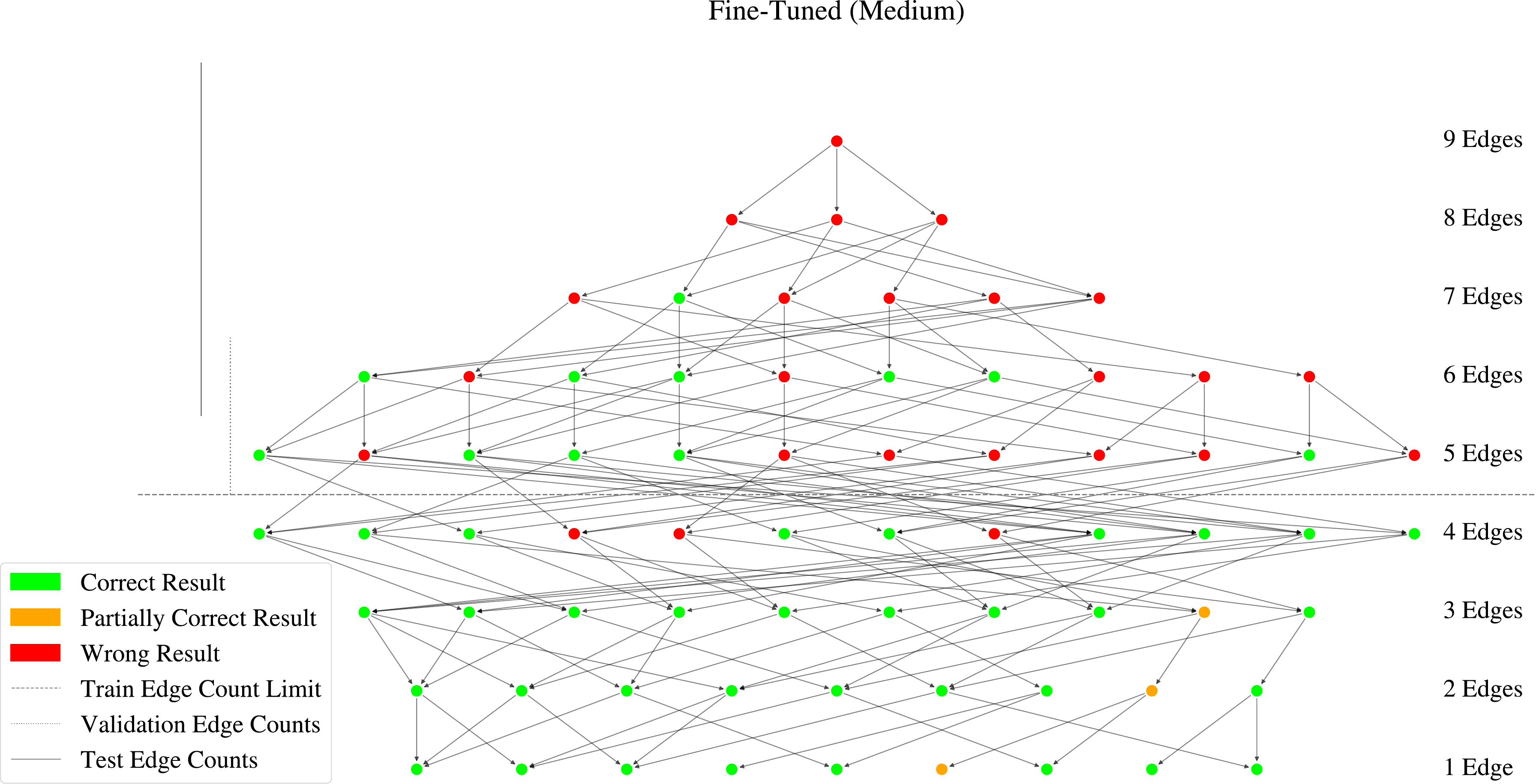}}
    \caption{Results for a depth and breadth $3$ graph pattern instance and connected sub-patterns taken from the best fine-tuned \emph{medium} approach. Arrows indicate source contains target node pattern, omitting nodes with $>1$ edge difference.}
    \label{fig:tree-medium-ft}
\end{figure}

Fig. \ref{fig:tree-medium-ft} shows all results for a single graph pattern instance of depth and breadth $3$ as well as the corresponding sub-graphs from the medium dataset for the best fine-tuned approach. The illustration can be understood similarly as a Hasse diagram \cite{hassediagram,hassediagram2} for the ``subset of'' relation between two graphs. The nodes in the bottom line correspond to single-triple questions, e.g., ``Who is the president of X?'', while the second-bottom line is one triple richer, e.g., ``Who is the father of the president of X?''. Thus, the figure illustrates the questions that can be built from the ``atomic'' building blocks up to the whole pattern instance. We can observe that data that were available during training plus up to two edges work relatively well, but almost all predictions beyond that are incorrect. Similarly, most non-training results violate the systematicity property.%

\section{Discussion}

Before we conclude to answer our research questions, we will first go through the results with the hypothesis that LLMs do work in a systematic (Def. \ref{def:sys}) manner and see whether this view is compatible with our empirical results.

\paragraph{Zero-Shot Experiments:}

In an ideal scenario, the models have already seen sufficient amounts of data for DBpedia during pre-training, such that it only needs a suitable explanation of the task to perform well. To ensure the explanation is not the problem here, we evaluated multiple prompt optimization techniques to find the best prompt for each tested model. However, the poor results (macro $F_1$ between $0.01$ and $0.04$) suggest that our assumptions are not true and the models need more information to successfully deal with QALD. %

\paragraph{Few-Shot Experiments:}

Thus, the natural next step when zero-shot prompting fails is to additionally provide examples/``shots'' to the model. 
However, we can observe in Table \ref{tab:results} as well as Fig. \ref{fig:heatmap-best-max} that, while it improves results, the absolute scores are still poor. Even for single-triple questions, the macro $F_1$ score is only $0.51$ in Fig. \ref{fig:heatmap-best-max-few}. The test compositionality scores are not much better either, with scores between $0.00$ and $0.07$. Similarly to the zero-shot experiments, the weak performance as well as the low compositionality scores question the hypothesis that LLMs work in a compositionally systematic manner.

\paragraph{Fine-Tuning Experiments:}

One reason that makes models struggle with the QALD task could be that, in order to construct a correct DBpedia SPARQL query for a specific question, one also needs to know how to deal with peculiarities in the mapping from natural language to DBpedia vocabulary.  The peculiarities could be learned by a model through fine-tuning, as the model then has the chance to learn to connect certain entities and properties to certain natural language expressions. As all entities, properties and ``atomic'' natural language expressions are by design part of the training data, ambiguity problems should be minimal after fine-tuning. %
Although the results of (classic) fine-tuning presented in Table \ref{tab:results} are much better overall, there are a few trends suggesting that the model has substantial problems to generalize beyond training data. First, the macro $F_1$ performance degrades to almost few-shot level as the deviation of samples from the training examples grows (i.e., for \emph{easy}, test performance jumps from $0.09$ to $0.45$, but for \emph{hard} only from $0.08$ to $0.09$). This impression is corroborated by the results presented in Fig. \ref{fig:heatmap-best-max-ft}, where the top left three rectangles, which represent the training data, achieve almost perfect results, whereas almost all other examples slightly outside the training data already display substantial performance drops. 
This impression that the models lack a proper generalization is further corroborated by Fig. \ref{fig:tree-medium-ft}. 
As we can see from the green nodes in the graph, most of the training split results are correct. However, this performance severely degrades when adding edges beyond the training data size. Thus, even when adding one additional edge, the number of respective incorrect results more than doubles and continues to grow. This indicates that the models struggle to generalize on a compositional level, contradicting our hypothesis.

\paragraph{Self-Contained Few-Shot and Fine-Tuning Experiments:}

Especially for the in-context learning experiments, 
we might discard the assumption that the models have already seen the relevant data from DBpedia such that we need to provide all necessary information in the input. This is what we did with the self-contained experiments, which contain an edge set cover of the target graph pattern, i.e., the respective questions and queries, in the input. Although this eases the task a lot and comes at the risk of overfitting during prompt optimization or fine-tuning,  
it also focuses the task on the core ability we want to test, namely systematicity. Thus, self-contained tasks provide almost optimal conditions for a model to show its abilities to work in a compositional way.

However, the performance improvements in Table \ref{tab:results} are not as large as one would expect assuming our initial hypothesis that the models indeed work in a compositional, systematic manner. The best-performing self-contained experiments in all cases outperform the classic experiments in terms of test macro $F_1$ scores and are less affected by the portion of edges given as training samples, i.e., the \emph{easy}, \emph{medium} and \emph{hard} datasets. This might be due to the fact that all relevant information is given in the input and always with (up to) $10$ shots, only with varying closeness of the given examples to the target question due to the edge portion limit. Notably, fine-tuning achieves worse results for the \emph{easy} and \emph{hard} datasets than the few-shot approach ($0.48$ vs. $0.57$ and $0.30$ vs. $0.51$). Only for \emph{medium} the fine-tuning approach works better ($0.55$ vs. $0.48$). It is not clear what this deviation is caused by. One possibility would be that the fine-tuning process overfits stronger to the training dataset and thus has more problems generalizing to larger target questions, whereas for in-context learning, only the prompt is optimized, with less potential to overfit. Regarding compositionality scores, we see better overall results as well, but with slight differences between approaches. For fine-tuning, the test compositionality $F_1$ score is in all cases slightly higher than the ``raw'' macro $F_1$ score, whereas it is lower for few-shot prompting. This might suggest that fine-tuning leads to better compositional behavior than few-shot prompting. Nevertheless, given the discussed simplifications and all scores being no greater than $0.57$ for both macro and compositionality $F_1$ score, this raises questions how compositionally LLMs are able to work and may indicate a fundamental weakness, especially considering the observations for higher depths, breadths, or edge counts. Overall, our results are in line with those in Dziri et al. \cite{allenaicomp}, indicating fundamental weaknesses in LLMs w.r.t. compositional reasoning.

\paragraph{Research Questions:} Considering \ref{rq1}, we presented an operationalization of compositionality in terms of systematicity in Def. \ref{def:sys} and Section \ref{sec:operationalization}. When it comes to whether or not LLMs satisfy the property defined in Def. \ref{def:sys} w.r.t. \ref{rq2}, our results indicate that the LLMs tested do not satisfy the property in a strict sense. Considering the merely mediocre scores even under optimal conditions, this at least heavily questions the general abilities to work in a compositional way w.r.t. the QALD domain. For \ref{rq3}, we have seen promising results that fine-tuning slightly improves the compositional behavior of LLMs, however, without achieving scores that would justify the assumption that LLMs actually satisfy systematicity in terms of Def. \ref{def:sys} when fine-tuned accordingly.

\subsection{Limitations}

Our study has multiple limitations that frame the scope and interpretation of our results. First, the dataset used is synthetic and constructed with a controlled set of lexical entries and structural templates. This therefore limits the diversity of surface verbalizations. However, we see this primarily as an advantage in the sense that by minimizing linguistic variation, we can reduce confounding factors such as paraphrasing and noise from formulation differences, thereby isolating the core ability of models to generalize compositionally. We aimed to address shortcut learning and overfitting by running diverse experiments with and without optimization/fine-tuning. If the limited linguistic diversity had a substantial impact on the (optimization) performance, this would have been visible in the results in one way or another. If the optimizations would hurt performance, this would usually be visible by experiments with non-optimized models performing better. If the optimization found an easy shortcut for the datasets, the scores would have been very high. All in all, both is not the case, such that we assume the simplicity to have a positive if any impact on the results. Secondly, it has to be highlighted that our dataset is built on a curated but ultimately limited subset of DBpedia. Although the coverage is relatively large, it does not represent the full heterogeneity of the DBpedia knowledge graph. Furthermore, DBpedia exhibits certain idiosyncrasies, particularly in its type definitions and predicate choices for particular information, that may result in generated benchmark items that appear to be ``unnatural''. Generalization to other knowledge bases, such as Wikidata, remains to be tested in future work. %
Additionally, we conducted only a limited set of fine-tuning experiments due to resource constraints. More experiments or even specialized training techniques for compositionality may yield further novel insights. Finally, we emphasize that our evaluation focuses on necessary conditions for compositionality. Success on our task does not definitively demonstrate that a model reasons compositionally in a systematic fashion. However, failure to generalize in our reduced and highly controlled setting strongly suggests a lack of compositional generalization capabilities.

\section{Conclusion}
In this paper, we presented the \emph{CompoST} dataset for testing systematic compositionality for the QALD task, together with benchmark results for a diverse set of current LLMs and optimization strategies. 
Our results suggest that LLMs face limitations when it comes to systematic compositional behavior, with increasing performance degradations the larger the deviation from the training data is. For example, the best-performing classic experiment achieves a test macro $F_1$ score of $0.45$ for the \emph{easy} dataset, but only $0.09$ for \emph{hard}. Even for self-contained experiments, with all necessary information in the input, only test macro $F_1$ scores between $0.30$ and $0.57$ are achieved. All in all, our results do not justify the assumption that LLMs work in a compositional manner. Future work could further investigate whether there are more specialized training techniques not covered by this paper which might improve their ability to work in a compositional manner. 

\paragraph*{Supplemental Material Statement:} Source code and dataset are available at \url{https://doi.org/10.5281/zenodo.16312287}.
\begin{credits}

\subsubsection{\ackname} This work is partially funded by the Ministry of Culture and Science of the State of North Rhine-Westphalia under grant no NW21-059A (SAIL).

This preprint has not undergone peer review (when applicable) or any post-submission improvements or corrections. The Version of Record of this contribution is published in The Semantic Web -- ISWC 2025, Lecture Notes in Computer Science (LNCS, volume 16140) and is available online at \url{https://doi.org/10.1007/978-3-032-09527-5_1}.

\subsubsection{\discintname}
The authors have no competing interests to declare that are relevant to the content of this article.
\end{credits}
\bibliographystyle{splncs04}
\bibliography{references}

\begin{thebibliography}{10}
\providecommand{\url}[1]{\texttt{#1}}
\providecommand{\urlprefix}{URL }
\providecommand{\doi}[1]{https://doi.org/#1}

\bibitem{phi4}
Abdin, M., Aneja, J., Behl, H., Bubeck, S., Eldan, R., Gunasekar, S., Harrison, M., Hewett, R.J., Javaheripi, M., Kauffmann, P., Lee, J.R., Lee, Y.T., Li, Y., Liu, W., Mendes, C.C.T., Nguyen, A., Price, E., Rosa, G.d., Saarikivi, O., Salim, A., Shah, S., Wang, X., Ward, R., Wu, Y., Yu, D., Zhang, C., Zhang, Y.: Phi-4 {Technical} {Report} (Dec 2024). \doi{10.48550/arXiv.2412.08905}, \url{http://arxiv.org/abs/2412.08905}, arXiv:2412.08905 [cs]

\bibitem{optuna}
Akiba, T., Sano, S., Yanase, T., Ohta, T., Koyama, M.: Optuna: a next-generation hyperparameter optimization framework. In: Proceedings of the 25th {ACM} {SIGKDD} international conference on knowledge discovery and data mining (2019)

\bibitem{bnf}
Backus, J.W.: The syntax and semantics of the proposed international algebraic language of the {Zurich} {ACM}-{GAMM} {Conference}. In: Information processing, proceedings of the 1st international conference on information processing, {UNESCO}, paris 15-20 june 1959. pp. 125--131. UNESCO (Paris) (1959), tex.bibsource: dblp computer science bibliography, https://dblp.org tex.timestamp: Fri, 26 Jul 2019 12:25:09 +0200

\bibitem{ageofllms}
Birhane, A., Kasirzadeh, A., Leslie, D., Wachter, S.: Science in the age of large language models. Nature Reviews Physics  \textbf{5}(5),  277--280 (May 2023). \doi{10.1038/s42254-023-00581-4}, \url{https://www.nature.com/articles/s42254-023-00581-4}, publisher: Nature Publishing Group

\bibitem{hassediagram}
Birkhoff, G.: Lattice {Theory}, revised ed. In: American mathematical society colloquium publications. vol.~25 (1948)

\bibitem{brown_language_2020}
Brown, T.B., Mann, B., Ryder, N., Subbiah, M., Kaplan, J., Dhariwal, P., Neelakantan, A., Shyam, P., Sastry, G., Askell, A., Agarwal, S., Herbert-Voss, A., Krueger, G., Henighan, T., Child, R., Ramesh, A., Ziegler, D.M., Wu, J., Winter, C., Hesse, C., Chen, M., Sigler, E., Litwin, M., Gray, S., Chess, B., Clark, J., Berner, C., McCandlish, S., Radford, A., Sutskever, I., Amodei, D.: Language {Models} are {Few}-{Shot} {Learners}. Advances in Neural Information Processing Systems  \textbf{33},  1877--1901 (2020)

\bibitem{chakraborty_multi-hop_2024}
Chakraborty, A.: Multi-hop {Question} {Answering} over {Knowledge} {Graphs} using {Large} {Language} {Models} (Apr 2024). \doi{10.48550/arXiv.2404.19234}, \url{http://arxiv.org/abs/2404.19234}, arXiv:2404.19234 [cs]

\bibitem{theocomp1}
Chen, L., Peng, B., Wu, H.: Theoretical limitations of multi-layer {Transformer} (2024), \url{https://arxiv.org/abs/2412.02975}, arXiv: 2412.02975 [cs.LG]

\bibitem{dessi_cnns_2019}
Dessì, R., Baroni, M.: {CNNs} found to jump around more skillfully than {RNNs}: {Compositional} {Generalization} in {Seq2seq} {Convolutional} {Networks}. In: Proceedings of the 57th {Annual} {Meeting} of the {Association} for {Computational} {Linguistics}. pp. 3919--3923. Association for Computational Linguistics, Florence, Italy (2019). \doi{10.18653/v1/P19-1381}, \url{https://www.aclweb.org/anthology/P19-1381}

\bibitem{devlin_bert_2019}
Devlin, J., Chang, M.W., Lee, K., Toutanova, K.: {BERT}: {Pre}-training of {Deep} {Bidirectional} {Transformers} for {Language} {Understanding} (May 2019). \doi{10.48550/arXiv.1810.04805}, \url{http://arxiv.org/abs/1810.04805}, arXiv:1810.04805 [cs]

\bibitem{ghidini_lc-quad_2019}
Dubey, M., Banerjee, D., Abdelkawi, A., Lehmann, J.: {LC}-{QuAD} 2.0: {A} {Large} {Dataset} for {Complex} {Question} {Answering} over {Wikidata} and {DBpedia}. In: Ghidini, C., Hartig, O., Maleshkova, M., Svátek, V., Cruz, I., Hogan, A., Song, J., Lefrançois, M., Gandon, F. (eds.) The {Semantic} {Web} – {ISWC} 2019, vol. 11779, pp. 69--78. Springer International Publishing, Cham (2019). \doi{10.1007/978-3-030-30796-7_5}, \url{https://link.springer.com/10.1007/978-3-030-30796-7_5}, series Title: Lecture Notes in Computer Science

\bibitem{allenaicomp}
Dziri, N., Lu, X., Sclar, M., Li, X.L., Jiang, L., Lin, B.Y., West, P., Bhagavatula, C., Le~Bras, R., Hwang, J.D., Sanyal, S., Welleck, S., Ren, X., Ettinger, A., Harchaoui, Z., Choi, Y.: Faith and fate: limits of transformers on compositionality. In: Proceedings of the 37th international conference on neural information processing systems. Nips '23, Curran Associates Inc., Red Hook, NY, USA (2023), number of pages: 40 Place: New Orleans, LA, USA tex.articleno: 3081

\bibitem{lightning}
Falcon, W., {The PyTorch Lightning team}: {PyTorch} lightning (Mar 2019). \doi{10.5281/zenodo.3828935}, \url{https://github.com/Lightning-AI/lightning}

\bibitem{theocomp3}
Feng, G., Zhang, B., Gu, Y., Ye, H., He, D., Wang, L.: Towards revealing the mystery behind chain of thought: a theoretical perspective. In: Oh, A., Naumann, T., Globerson, A., Saenko, K., Hardt, M., Levine, S. (eds.) Advances in neural information processing systems. vol.~36, pp. 70757--70798. Curran Associates, Inc. (2023), \url{https://proceedings.neurips.cc/paper_files/paper/2023/file/dfc310e81992d2e4cedc09ac47eff13e-Paper-Conference.pdf}

\bibitem{furrer_compositional_2021}
Furrer, D., Zee, M.v., Scales, N., Schärli, N.: Compositional {Generalization} in {Semantic} {Parsing}: {Pre}-training vs. {Specialized} {Architectures} (Sep 2021). \doi{10.48550/arXiv.2007.08970}, \url{http://arxiv.org/abs/2007.08970}, arXiv:2007.08970 [cs]

\bibitem{llama}
Grattafiori, A., Dubey, A., Jauhri, A., Pandey, A., Kadian, A., Al-Dahle, A., Letman, A., Mathur, A., Schelten, A., Vaughan, A., Yang, A., Fan, A., Goyal, A., Hartshorn, A., Yang, A., Mitra, A., Sravankumar, A., Korenev, A., Hinsvark, A., Rao, A., Zhang, A., Rodriguez, A., Gregerson, A., Spataru, A., Roziere, B., Biron, B., Tang, B., Chern, B., Caucheteux, C., Nayak, C., Bi, C., Marra, C., McConnell, C., Keller, C., Touret, C., Wu, C., Wong, C., Ferrer, C.C., Nikolaidis, C., Allonsius, D., Song, D., Pintz, D., Livshits, D., Wyatt, D., Esiobu, D., Choudhary, D., Mahajan, D., Garcia-Olano, D., Perino, D., Hupkes, D., Lakomkin, E., AlBadawy, E., Lobanova, E., Dinan, E., Smith, E.M., Radenovic, F., Guzmán, F., Zhang, F., Synnaeve, G., Lee, G., Anderson, G.L., Thattai, G., Nail, G., Mialon, G., Pang, G., Cucurell, G., Nguyen, H., Korevaar, H., Xu, H., Touvron, H., Zarov, I., Ibarra, I.A., Kloumann, I., Misra, I., Evtimov, I., Zhang, J., Copet, J., Lee, J., Geffert, J., Vranes, J., Park, J., Mahadeokar, J., Shah,
  J., Linde, J.v.d., Billock, J., Hong, J., Lee, J., Fu, J., Chi, J., Huang, J., Liu, J., Wang, J., Yu, J., Bitton, J., Spisak, J., Park, J., Rocca, J., Johnstun, J., Saxe, J., Jia, J., Alwala, K.V., Prasad, K., Upasani, K., Plawiak, K., Li, K., Heafield, K., Stone, K., El-Arini, K., Iyer, K., Malik, K., Chiu, K., Bhalla, K., Lakhotia, K., Rantala-Yeary, L., Maaten, L.v.d., Chen, L., Tan, L., Jenkins, L., Martin, L., Madaan, L., Malo, L., Blecher, L., Landzaat, L., Oliveira, L.d., Muzzi, M., Pasupuleti, M., Singh, M., Paluri, M., Kardas, M., Tsimpoukelli, M., Oldham, M., Rita, M., Pavlova, M., Kambadur, M., Lewis, M., Si, M., Singh, M.K., Hassan, M., Goyal, N., Torabi, N., Bashlykov, N., Bogoychev, N., Chatterji, N., Zhang, N., Duchenne, O., Çelebi, O., Alrassy, P., Zhang, P., Li, P., Vasic, P., Weng, P., Bhargava, P., Dubal, P., Krishnan, P., Koura, P.S., Xu, P., He, Q., Dong, Q., Srinivasan, R., Ganapathy, R., Calderer, R., Cabral, R.S., Stojnic, R., Raileanu, R., Maheswari, R., Girdhar, R., Patel, R.,
  Sauvestre, R., Polidoro, R., Sumbaly, R., Taylor, R., Silva, R., Hou, R., Wang, R., Hosseini, S., Chennabasappa, S., Singh, S., Bell, S., Kim, S.S., Edunov, S., Nie, S., Narang, S., Raparthy, S., Shen, S., Wan, S., Bhosale, S., Zhang, S., Vandenhende, S., Batra, S., Whitman, S., Sootla, S., Collot, S., Gururangan, S., Borodinsky, S., Herman, T., Fowler, T., Sheasha, T., Georgiou, T., Scialom, T., Speckbacher, T., Mihaylov, T., Xiao, T., Karn, U., Goswami, V., Gupta, V., Ramanathan, V., Kerkez, V., Gonguet, V., Do, V., Vogeti, V., Albiero, V., Petrovic, V., Chu, W., Xiong, W., Fu, W., Meers, W., Martinet, X., Wang, X., Wang, X., Tan, X.E., Xia, X., Xie, X., Jia, X., Wang, X., Goldschlag, Y., Gaur, Y., Babaei, Y., Wen, Y., Song, Y., Zhang, Y., Li, Y., Mao, Y., Coudert, Z.D., Yan, Z., Chen, Z., Papakipos, Z., Singh, A., Srivastava, A., Jain, A., Kelsey, A., Shajnfeld, A., Gangidi, A., Victoria, A., Goldstand, A., Menon, A., Sharma, A., Boesenberg, A., Baevski, A., Feinstein, A., Kallet, A., Sangani, A., Teo,
  A., Yunus, A., Lupu, A., Alvarado, A., Caples, A., Gu, A., Ho, A., Poulton, A., Ryan, A., Ramchandani, A., Dong, A., Franco, A., Goyal, A., Saraf, A., Chowdhury, A., Gabriel, A., Bharambe, A., Eisenman, A., Yazdan, A., James, B., Maurer, B., Leonhardi, B., Huang, B., Loyd, B., Paola, B.D., Paranjape, B., Liu, B., Wu, B., Ni, B., Hancock, B., Wasti, B., Spence, B., Stojkovic, B., Gamido, B., Montalvo, B., Parker, C., Burton, C., Mejia, C., Liu, C., Wang, C., Kim, C., Zhou, C., Hu, C., Chu, C.H., Cai, C., Tindal, C., Feichtenhofer, C., Gao, C., Civin, D., Beaty, D., Kreymer, D., Li, D., Adkins, D., Xu, D., Testuggine, D., David, D., Parikh, D., Liskovich, D., Foss, D., Wang, D., Le, D., Holland, D., Dowling, E., Jamil, E., Montgomery, E., Presani, E., Hahn, E., Wood, E., Le, E.T., Brinkman, E., Arcaute, E., Dunbar, E., Smothers, E., Sun, F., Kreuk, F., Tian, F., Kokkinos, F., Ozgenel, F., Caggioni, F., Kanayet, F., Seide, F., Florez, G.M., Schwarz, G., Badeer, G., Swee, G., Halpern, G., Herman, G., Sizov, G.,
  Guangyi, Zhang, Lakshminarayanan, G., Inan, H., Shojanazeri, H., Zou, H., Wang, H., Zha, H., Habeeb, H., Rudolph, H., Suk, H., Aspegren, H., Goldman, H., Zhan, H., Damlaj, I., Molybog, I., Tufanov, I., Leontiadis, I., Veliche, I.E., Gat, I., Weissman, J., Geboski, J., Kohli, J., Lam, J., Asher, J., Gaya, J.B., Marcus, J., Tang, J., Chan, J., Zhen, J., Reizenstein, J., Teboul, J., Zhong, J., Jin, J., Yang, J., Cummings, J., Carvill, J., Shepard, J., McPhie, J., Torres, J., Ginsburg, J., Wang, J., Wu, K., U, K.H., Saxena, K., Khandelwal, K., Zand, K., Matosich, K., Veeraraghavan, K., Michelena, K., Li, K., Jagadeesh, K., Huang, K., Chawla, K., Huang, K., Chen, L., Garg, L., A, L., Silva, L., Bell, L., Zhang, L., Guo, L., Yu, L., Moshkovich, L., Wehrstedt, L., Khabsa, M., Avalani, M., Bhatt, M., Mankus, M., Hasson, M., Lennie, M., Reso, M., Groshev, M., Naumov, M., Lathi, M., Keneally, M., Liu, M., Seltzer, M.L., Valko, M., Restrepo, M., Patel, M., Vyatskov, M., Samvelyan, M., Clark, M., Macey, M., Wang, M.,
  Hermoso, M.J., Metanat, M., Rastegari, M., Bansal, M., Santhanam, N., Parks, N., White, N., Bawa, N., Singhal, N., Egebo, N., Usunier, N., Mehta, N., Laptev, N.P., Dong, N., Cheng, N., Chernoguz, O., Hart, O., Salpekar, O., Kalinli, O., Kent, P., Parekh, P., Saab, P., Balaji, P., Rittner, P., Bontrager, P., Roux, P., Dollar, P., Zvyagina, P., Ratanchandani, P., Yuvraj, P., Liang, Q., Alao, R., Rodriguez, R., Ayub, R., Murthy, R., Nayani, R., Mitra, R., Parthasarathy, R., Li, R., Hogan, R., Battey, R., Wang, R., Howes, R., Rinott, R., Mehta, S., Siby, S., Bondu, S.J., Datta, S., Chugh, S., Hunt, S., Dhillon, S., Sidorov, S., Pan, S., Mahajan, S., Verma, S., Yamamoto, S., Ramaswamy, S., Lindsay, S., Lindsay, S., Feng, S., Lin, S., Zha, S.C., Patil, S., Shankar, S., Zhang, S., Zhang, S., Wang, S., Agarwal, S., Sajuyigbe, S., Chintala, S., Max, S., Chen, S., Kehoe, S., Satterfield, S., Govindaprasad, S., Gupta, S., Deng, S., Cho, S., Virk, S., Subramanian, S., Choudhury, S., Goldman, S., Remez, T., Glaser, T.,
  Best, T., Koehler, T., Robinson, T., Li, T., Zhang, T., Matthews, T., Chou, T., Shaked, T., Vontimitta, V., Ajayi, V., Montanez, V., Mohan, V., Kumar, V.S., Mangla, V., Ionescu, V., Poenaru, V., Mihailescu, V.T., Ivanov, V., Li, W., Wang, W., Jiang, W., Bouaziz, W., Constable, W., Tang, X., Wu, X., Wang, X., Wu, X., Gao, X., Kleinman, Y., Chen, Y., Hu, Y., Jia, Y., Qi, Y., Li, Y., Zhang, Y., Zhang, Y., Adi, Y., Nam, Y., Yu, Wang, Zhao, Y., Hao, Y., Qian, Y., Li, Y., He, Y., Rait, Z., DeVito, Z., Rosnbrick, Z., Wen, Z., Yang, Z., Zhao, Z., Ma, Z.: The {Llama} 3 {Herd} of {Models} (Nov 2024). \doi{10.48550/arXiv.2407.21783}, \url{http://arxiv.org/abs/2407.21783}, arXiv:2407.21783 [cs]

\bibitem{he_mintqa_2025}
He, J., Hu, N., Long, W., Chen, J., Pan, J.Z.: {MINTQA}: {A} {Multi}-{Hop} {Question} {Answering} {Benchmark} for {Evaluating} {LLMs} on {New} and {Tail} {Knowledge} (Jan 2025). \doi{10.48550/arXiv.2412.17032}, \url{http://arxiv.org/abs/2412.17032}, arXiv:2412.17032 [cs]

\bibitem{herzig_unlocking_2021}
Herzig, J., Shaw, P., Chang, M.W., Guu, K., Pasupat, P., Zhang, Y.: Unlocking {Compositional} {Generalization} in {Pre}-trained {Models} {Using} {Intermediate} {Representations} (Apr 2021). \doi{10.48550/arXiv.2104.07478}, \url{http://arxiv.org/abs/2104.07478}, arXiv:2104.07478 [cs]

\bibitem{qwen2}
Hui, B., Yang, J., Cui, Z., Yang, J., Liu, D., Zhang, L., Liu, T., Zhang, J., Yu, B., Dang, K., {others}: Qwen2. 5-coder technical report. arXiv preprint arXiv:2409.12186  (2024)

\bibitem{hupkes_compositionality_2020}
Hupkes, D., Dankers, V., Mul, M., Bruni, E.: Compositionality {Decomposed}: {How} do {Neural} {Networks} {Generalise}? Journal of Artificial Intelligence Research  \textbf{67},  757--795 (Apr 2020). \doi{10.1613/jair.1.11674}, \url{https://jair.org/index.php/jair/article/view/11674}

\bibitem{ismayilzada_evaluating_2025}
Ismayilzada, M., Circi, D., Sälevä, J., Sirin, H., Köksal, A., Dhingra, B., Bosselut, A., Ataman, D., Plas, L.v.d.: Evaluating {Morphological} {Compositional} {Generalization} in {Large} {Language} {Models} (Feb 2025). \doi{10.48550/arXiv.2410.12656}, \url{http://arxiv.org/abs/2410.12656}, arXiv:2410.12656 [cs]

\bibitem{keysers_measuring_2020}
Keysers, D., Schärli, N., Scales, N., Buisman, H., Furrer, D., Kashubin, S., Momchev, N., Sinopalnikov, D., Stafiniak, L., Tihon, T., Tsarkov, D., Wang, X., Zee, M.v., Bousquet, O.: Measuring {Compositional} {Generalization}: {A} {Comprehensive} {Method} on {Realistic} {Data} (Jun 2020). \doi{10.48550/arXiv.1912.09713}, \url{http://arxiv.org/abs/1912.09713}, arXiv:1912.09713 [cs]

\bibitem{dspy2}
Khattab, O., Santhanam, K., Li, X.L., Hall, D., Liang, P., Potts, C., Zaharia, M.: Demonstrate-search-predict: {Composing} retrieval and language models for knowledge-intensive {NLP}. arXiv preprint arXiv:2212.14024  (2022)

\bibitem{dspy1}
Khattab, O., Singhvi, A., Maheshwari, P., Zhang, Z., Santhanam, K., Vardhamanan, S., Haq, S., Sharma, A., Joshi, T.T., Moazam, H., Miller, H., Zaharia, M., Potts, C.: {DSPy}: {Compiling} declarative language model calls into self-improving pipelines. The {Twelfth} {International} {Conference} on {Learning} {Representations} (2024)

\bibitem{kim_cogs_2020}
Kim, N., Linzen, T.: {COGS}: {A} {Compositional} {Generalization} {Challenge} {Based} on {Semantic} {Interpretation} (Oct 2020). \doi{10.48550/arXiv.2010.05465}, \url{http://arxiv.org/abs/2010.05465}, arXiv:2010.05465 [cs]

\bibitem{lake_generalization_2018}
Lake, B., Baroni, M.: Generalization without {Systematicity}:{On} the {Compositional} {Skills} of {Sequence}-to-{Sequence} {Recurrent} {Networks}  (2018)

\bibitem{lemon}
McCrae, J.P., Spohr, D., Cimiano, P.: Linking lexical resources and ontologies on the semantic web with lemon. In: Proceedings of the 8th extended semantic web conference on {The} semantic web: research and applications ({ESWC}). vol.~6643, pp. 245--259 (2011)

\bibitem{theocomp2}
McLeish, S., Bansal, A., Stein, A., Jain, N., Kirchenbauer, J., Bartoldson, B.R., Kailkhura, B., Bhatele, A., Geiping, J., Schwarzschild, A., Goldstein, T.: Transformers can do arithmetic with the right embeddings. In: Globerson, A., Mackey, L., Belgrave, D., Fan, A., Paquet, U., Tomczak, J., Zhang, C. (eds.) Advances in neural information processing systems. vol.~37, pp. 108012--108041. Curran Associates, Inc. (2024), \url{https://proceedings.neurips.cc/paper_files/paper/2024/file/c35986bc1ee29b31c1011481b77fe540-Paper-Conference.pdf}

\bibitem{theocomp5}
Merrill, W., Sabharwal, A.: The parallelism tradeoff: {Limitations} of log-precision transformers. Transactions of the Association for Computational Linguistics  \textbf{11},  531--545 (Jun 2023). \doi{10.1162/tacl_a_00562}, \url{https://doi.org/10.1162/tacl_a_00562}, tex.eprint: https://direct.mit.edu/tacl/article-pdf/doi/10.1162/tacl{\textbackslash}\_a{\textbackslash}\_00562/2131191/tacl{\textbackslash}\_a{\textbackslash}\_00562.pdf

\bibitem{nye_show_2021}
Nye, M., Andreassen, A.J., Gur-Ari, G., Michalewski, H., Austin, J., Bieber, D., Dohan, D., Lewkowycz, A., Bosma, M., Luan, D., Sutton, C., Odena, A.: Show {Your} {Work}: {Scratchpads} for {Intermediate} {Computation} with {Language} {Models} (Nov 2021). \doi{10.48550/arXiv.2112.00114}, \url{http://arxiv.org/abs/2112.00114}, arXiv:2112.00114 [cs]

\bibitem{olmo}
OLMo, T., Walsh, P., Soldaini, L., Groeneveld, D., Lo, K., Arora, S., Bhagia, A., Gu, Y., Huang, S., Jordan, M., Lambert, N., Schwenk, D., Tafjord, O., Anderson, T., Atkinson, D., Brahman, F., Clark, C., Dasigi, P., Dziri, N., Guerquin, M., Ivison, H., Koh, P.W., Liu, J., Malik, S., Merrill, W., Miranda, L.J.V., Morrison, J., Murray, T., Nam, C., Pyatkin, V., Rangapur, A., Schmitz, M., Skjonsberg, S., Wadden, D., Wilhelm, C., Wilson, M., Zettlemoyer, L., Farhadi, A., Smith, N.A., Hajishirzi, H.: 2 {OLMo} 2 {Furious} (Jan 2025). \doi{10.48550/arXiv.2501.00656}, \url{http://arxiv.org/abs/2501.00656}, arXiv:2501.00656 [cs]

\bibitem{gpt4}
OpenAI, Achiam, J., Adler, S., Agarwal, S., Ahmad, L., Akkaya, I., Aleman, F.L., Almeida, D., Altenschmidt, J., Altman, S., Anadkat, S., Avila, R., Babuschkin, I., Balaji, S., Balcom, V., Baltescu, P., Bao, H., Bavarian, M., Belgum, J., Bello, I., Berdine, J., Bernadett-Shapiro, G., Berner, C., Bogdonoff, L., Boiko, O., Boyd, M., Brakman, A.L., Brockman, G., Brooks, T., Brundage, M., Button, K., Cai, T., Campbell, R., Cann, A., Carey, B., Carlson, C., Carmichael, R., Chan, B., Chang, C., Chantzis, F., Chen, D., Chen, S., Chen, R., Chen, J., Chen, M., Chess, B., Cho, C., Chu, C., Chung, H.W., Cummings, D., Currier, J., Dai, Y., Decareaux, C., Degry, T., Deutsch, N., Deville, D., Dhar, A., Dohan, D., Dowling, S., Dunning, S., Ecoffet, A., Eleti, A., Eloundou, T., Farhi, D., Fedus, L., Felix, N., Fishman, S.P., Forte, J., Fulford, I., Gao, L., Georges, E., Gibson, C., Goel, V., Gogineni, T., Goh, G., Gontijo-Lopes, R., Gordon, J., Grafstein, M., Gray, S., Greene, R., Gross, J., Gu, S.S., Guo, Y., Hallacy, C.,
  Han, J., Harris, J., He, Y., Heaton, M., Heidecke, J., Hesse, C., Hickey, A., Hickey, W., Hoeschele, P., Houghton, B., Hsu, K., Hu, S., Hu, X., Huizinga, J., Jain, S., Jain, S., Jang, J., Jiang, A., Jiang, R., Jin, H., Jin, D., Jomoto, S., Jonn, B., Jun, H., Kaftan, T., Kaiser, L., Kamali, A., Kanitscheider, I., Keskar, N.S., Khan, T., Kilpatrick, L., Kim, J.W., Kim, C., Kim, Y., Kirchner, J.H., Kiros, J., Knight, M., Kokotajlo, D., Kondraciuk, L., Kondrich, A., Konstantinidis, A., Kosic, K., Krueger, G., Kuo, V., Lampe, M., Lan, I., Lee, T., Leike, J., Leung, J., Levy, D., Li, C.M., Lim, R., Lin, M., Lin, S., Litwin, M., Lopez, T., Lowe, R., Lue, P., Makanju, A., Malfacini, K., Manning, S., Markov, T., Markovski, Y., Martin, B., Mayer, K., Mayne, A., McGrew, B., McKinney, S.M., McLeavey, C., McMillan, P., McNeil, J., Medina, D., Mehta, A., Menick, J., Metz, L., Mishchenko, A., Mishkin, P., Monaco, V., Morikawa, E., Mossing, D., Mu, T., Murati, M., Murk, O., Mély, D., Nair, A., Nakano, R., Nayak, R.,
  Neelakantan, A., Ngo, R., Noh, H., Ouyang, L., O'Keefe, C., Pachocki, J., Paino, A., Palermo, J., Pantuliano, A., Parascandolo, G., Parish, J., Parparita, E., Passos, A., Pavlov, M., Peng, A., Perelman, A., Peres, F.d.A.B., Petrov, M., Pinto, H.P.d.O., Michael, Pokorny, Pokrass, M., Pong, V.H., Powell, T., Power, A., Power, B., Proehl, E., Puri, R., Radford, A., Rae, J., Ramesh, A., Raymond, C., Real, F., Rimbach, K., Ross, C., Rotsted, B., Roussez, H., Ryder, N., Saltarelli, M., Sanders, T., Santurkar, S., Sastry, G., Schmidt, H., Schnurr, D., Schulman, J., Selsam, D., Sheppard, K., Sherbakov, T., Shieh, J., Shoker, S., Shyam, P., Sidor, S., Sigler, E., Simens, M., Sitkin, J., Slama, K., Sohl, I., Sokolowsky, B., Song, Y., Staudacher, N., Such, F.P., Summers, N., Sutskever, I., Tang, J., Tezak, N., Thompson, M.B., Tillet, P., Tootoonchian, A., Tseng, E., Tuggle, P., Turley, N., Tworek, J., Uribe, J.F.C., Vallone, A., Vijayvergiya, A., Voss, C., Wainwright, C., Wang, J.J., Wang, A., Wang, B., Ward, J., Wei,
  J., Weinmann, C.J., Welihinda, A., Welinder, P., Weng, J., Weng, L., Wiethoff, M., Willner, D., Winter, C., Wolrich, S., Wong, H., Workman, L., Wu, S., Wu, J., Wu, M., Xiao, K., Xu, T., Yoo, S., Yu, K., Yuan, Q., Zaremba, W., Zellers, R., Zhang, C., Zhang, M., Zhao, S., Zheng, T., Zhuang, J., Zhuk, W., Zoph, B.: {GPT}-4 {Technical} {Report} (Mar 2024). \doi{10.48550/arXiv.2303.08774}, \url{http://arxiv.org/abs/2303.08774}, arXiv:2303.08774 [cs]

\bibitem{panda_holmes_2024}
Panda, P., Agarwal, A., Devaguptapu, C., Kaul, M., Ap, P.: {HOLMES}: {Hyper}-{Relational} {Knowledge} {Graphs} for {Multi}-hop {Question} {Answering} using {LLMs}. In: Proceedings of the 62nd {Annual} {Meeting} of the {Association} for {Computational} {Linguistics} ({Volume} 1: {Long} {Papers}). pp. 13263--13282. Association for Computational Linguistics, Bangkok, Thailand (2024). \doi{10.18653/v1/2024.acl-long.717}, \url{https://aclanthology.org/2024.acl-long.717}

\bibitem{qald9plus}
Perevalov, A., Diefenbach, D., Usbeck, R., Both, A.: {QALD}-9-plus: {A} {Multilingual} {Dataset} for {Question} {Answering} over {DBpedia} and {Wikidata} {Translated} by {Native} {Speakers}. pp. 229--234. IEEE Computer Society (Jan 2022). \doi{10.1109/ICSC52841.2022.00045}, \url{https://www.computer.org/csdl/proceedings-article/icsc/2022/341800a229/1BYIptAeqty}, iSSN: 2325-6516

\bibitem{petty_impact_2024}
Petty, J., Steenkiste, S., Dasgupta, I., Sha, F., Garrette, D., Linzen, T.: The {Impact} of {Depth} on {Compositional} {Generalization} in {Transformer} {Language} {Models}. In: Duh, K., Gomez, H., Bethard, S. (eds.) Proceedings of the 2024 {Conference} of the {North} {American} {Chapter} of the {Association} for {Computational} {Linguistics}: {Human} {Language} {Technologies} ({Volume} 1: {Long} {Papers}). pp. 7239--7252. Association for Computational Linguistics, Mexico City, Mexico (Jun 2024). \doi{10.18653/v1/2024.naacl-long.402}, \url{https://aclanthology.org/2024.naacl-long.402/}

\bibitem{piantadosi_meaning_2022}
Piantadosi, S.T., Hill, F.: Meaning without reference in large language models (Aug 2022). \doi{10.48550/arXiv.2208.02957}, \url{http://arxiv.org/abs/2208.02957}, arXiv:2208.02957 [cs]

\bibitem{compgap}
Press, O., Zhang, M., Min, S., Schmidt, L., Smith, N., Lewis, M.: Measuring and {Narrowing} the {Compositionality} {Gap} in {Language} {Models}. In: Bouamor, H., Pino, J., Bali, K. (eds.) Findings of the {Association} for {Computational} {Linguistics}: {EMNLP} 2023. pp. 5687--5711. Association for Computational Linguistics, Singapore (Dec 2023). \doi{10.18653/v1/2023.findings-emnlp.378}, \url{https://aclanthology.org/2023.findings-emnlp.378/}

\bibitem{deepspeed}
Rasley, J., Rajbhandari, S., Ruwase, O., He, Y.: {DeepSpeed}: {System} {Optimizations} {Enable} {Training} {Deep} {Learning} {Models} with {Over} 100 {Billion} {Parameters}. In: Proceedings of the 26th {ACM} {SIGKDD} {International} {Conference} on {Knowledge} {Discovery} \& {Data} {Mining}. pp. 3505--3506. {KDD} '20, Association for Computing Machinery, New York, NY, USA (Aug 2020). \doi{10.1145/3394486.3406703}, \url{https://dl.acm.org/doi/10.1145/3394486.3406703}

\bibitem{hassediagram2}
Rival, I.: The {Diagram}. In: Rival, I. (ed.) Graphs and {Order}: {The} {Role} of {Graphs} in the {Theory} of {Ordered} {Sets} and {Its} {Applications}, pp. 103--133. Springer Netherlands, Dordrecht (1985). \doi{10.1007/978-94-009-5315-4_3}, \url{https://doi.org/10.1007/978-94-009-5315-4_3}

\bibitem{saparov_language_2023}
Saparov, A., He, H.: Language {Models} {Are} {Greedy} {Reasoners}: {A} {Systematic} {Formal} {Analysis} of {Chain}-of-{Thought} (Mar 2023). \doi{10.48550/arXiv.2210.01240}, \url{http://arxiv.org/abs/2210.01240}, arXiv:2210.01240 [cs]

\bibitem{neodudesPoster}
Schmidt, D.M., Elahi, M.F., Cimiano, P.: Lexicalization {Is} {All} {You} {Need}: {Examining} the {Impact} of {Lexical} {Knowledge} in a {Compositional} {QALD} {System}. In: Badenes-Olmedo, C., Novalija, I., Daga, E., Stork, L., Pillai, R.G., Dierickx, L., Kruit, B., Degeler, V., Moreira, J., Zhang, B., Alharbi, R., He, Y., Graciotti, A., Tirado, A.M., Presutti, V., Motta, E. (eds.) Joint {Proceedings} of {Posters}, {Demos}, {Workshops}, and {Tutorials} of the 24th {International} {Conference} on {Knowledge} {Engineering} and {Knowledge} {Management} ({EKAW}-{PDWT} 2024). {CEUR} {Workshop} {Proceedings}, vol.~3967. CEUR, Amsterdam, Netherlands (Nov 2024), \url{https://ceur-ws.org/Vol-3967/#PD_paper_160}, iSSN: 1613-0073

\bibitem{neodudes}
Schmidt, D.M., Elahi, M.F., Cimiano, P.: Lexicalization is all you need: {Examining} the impact of lexical knowledge in a compositional {QALD} system. In: Alam, M., Rospocher, M., van Erp, M., Hollink, L., Gesese, G.A. (eds.) Knowledge engineering and knowledge management. pp. 102--122. Springer Nature Switzerland, Cham (2025)

\bibitem{shekarpour2016question}
Shekarpour, S., Endris, K.M., Jaya~Kumar, A., Lukovnikov, D., Singh, K., Thakkar, H., Lange, C.: Question answering on linked data: {Challenges} and future directions. In: Proceedings of the 25th international conference companion on world wide web ({WWW}). pp. 693--698 (2016)

\bibitem{theocomp6}
Strobl, L., Merrill, W., Weiss, G., Chiang, D., Angluin, D.: What formal languages can transformers express? {A} survey. Transactions of the Association for Computational Linguistics  \textbf{12},  543--561 (2024). \doi{10.1162/tacl_a_00663}, \url{http://dx.doi.org/10.1162/tacl_a_00663}, publisher: MIT Press

\bibitem{comp}
Szabó, Z.G.: The case for compositionality. In: Werning, M., Hinzen, W., Machery, E. (eds.) The oxford handbook of compositionality. Oxford University Press (2012)

\bibitem{qald9}
Usbeck, R., Gusmita, R.H., Ngomo, A.C.N., Saleem, M.: 9th challenge on question answering over linked data ({QALD}-9). In: Joint proceedings of the 4th workshop on semantic deep learning ({SemDeep}-4) and {NLIWoD4}: {Natural} language interfaces for the web of data ({NLIWOD}-4) and 9th question answering over linked data challenge ({QALD}-9) co-located with 17th international semantic web conference ({ISWC}). pp. 58--64. California, United States of America (2018)

\bibitem{qald10}
Usbeck, R., Yan, X., Perevalov, A., Jiang, L., Schulz, J., Kraft, A., Möller, C., Huang, J., Reineke, J., Ngomo, A.C.N., Saleem, M., Both, A.: {QALD}-10 — the 10th challenge on question answering over linked data. Semantic Web  (Feb 2023), publisher: IOS Press BV

\bibitem{wei_chain--thought_2022}
Wei, J., Wang, X., Schuurmans, D., Bosma, M., Ichter, B., Xia, F., Chi, E.H., Le, Q.V., Zhou, D.: Chain-of-thought prompting elicits reasoning in large language models. In: Proceedings of the 36th {International} {Conference} on {Neural} {Information} {Processing} {Systems}. pp. 24824--24837. {NIPS} '22, Curran Associates Inc., Red Hook, NY, USA (Nov 2022)

\bibitem{welleck_naturalprover_2022}
Welleck, S., Liu, J., Lu, X., Hajishirzi, H., Choi, Y.: {NaturalProver}: {Grounded} {Mathematical} {Proof} {Generation} with {Language} {Models}. Advances in Neural Information Processing Systems  \textbf{35},  4913--4927 (Dec 2022), \url{https://proceedings.neurips.cc/paper_files/paper/2022/hash/1fc548a8243ad06616eee731e0572927-Abstract-Conference.html}

\bibitem{qwen1}
Yang, A., Yang, B., Hui, B., Zheng, B., Yu, B., Zhou, C., Li, C., Li, C., Liu, D., Huang, F., {others}: Qwen2 technical report. arXiv preprint arXiv:2407.10671  (2024)

\bibitem{yang_exploring_2024}
Yang, H., Lu, H., Lam, W., Cai, D.: Exploring {Compositional} {Generalization} of {Large} {Language} {Models}. In: Proceedings of the 2024 {Conference} of the {North} {American} {Chapter} of the {Association} for {Computational} {Linguistics}: {Human} {Language} {Technologies} ({Volume} 4: {Student} {Research} {Workshop}). pp. 16--24. Association for Computational Linguistics, Mexico City, Mexico (2024). \doi{10.18653/v1/2024.naacl-srw.3}, \url{https://aclanthology.org/2024.naacl-srw.3}

\bibitem{yih_semantic_2015}
Yih, W.t., Chang, M.W., He, X., Gao, J.: Semantic {Parsing} via {Staged} {Query} {Graph} {Generation}: {Question} {Answering} with {Knowledge} {Base}. In: Proceedings of the 53rd {Annual} {Meeting} of the {Association} for {Computational} {Linguistics} and the 7th {International} {Joint} {Conference} on {Natural} {Language} {Processing} ({Volume} 1: {Long} {Papers}). pp. 1321--1331. Association for Computational Linguistics, Beijing, China (2015). \doi{10.3115/v1/P15-1128}, \url{http://aclweb.org/anthology/P15-1128}

\bibitem{zhang_unveiling_2023}
Zhang, Y., Backurs, A., Bubeck, S., Eldan, R., Gunasekar, S., Wagner, T.: Unveiling {Transformers} with {LEGO}: a synthetic reasoning task (Feb 2023). \doi{10.48550/arXiv.2206.04301}, \url{http://arxiv.org/abs/2206.04301}, arXiv:2206.04301 [cs]

\bibitem{zhao_exploring_2024}
Zhao, J., Tong, J., Mou, Y., Zhang, M., Zhang, Q., Huang, X.: Exploring the {Compositional} {Deficiency} of {Large} {Language} {Models} in {Mathematical} {Reasoning} {Through} {Trap} {Problems}. In: Proceedings of the 2024 {Conference} on {Empirical} {Methods} in {Natural} {Language} {Processing}. pp. 16361--16376. Association for Computational Linguistics, Miami, Florida, USA (2024). \doi{10.18653/v1/2024.emnlp-main.915}, \url{https://aclanthology.org/2024.emnlp-main.915}

\bibitem{theocomp4}
Zubić, N., Soldá, F., Sulser, A., Scaramuzza, D.: Limits of deep learning: {Sequence} modeling through the lens of complexity theory (2025), \url{https://arxiv.org/abs/2405.16674}, arXiv: 2405.16674 [cs.LG]

\end{thebibliography}
\end{document}